\documentclass[runningheads]{llncs}
\usepackage{todonotes}
\usepackage{subfigure}
\usepackage{textgreek}
\usepackage{graphicx}
\usepackage{float}
\usepackage{wrapfig}
\usepackage{tabularx}
\usepackage[acronym]{glossaries}
\newacronym{ALS}{ALS}{airborne laser scanning}
\newacronym{MLS}{MLS}{mobile laser scanning}
\newacronym{LoD}{LoD}{level of detail}
\newacronym{OGC}{OGC}{Open Geospatial Consortium}
\newacronym{GML}{GML}{Geography Markup Language}
\newacronym{ASAM}{ASAM}{Association for Standardization of Automation and Measuring Systems}
\newacronym{TLS}{TLS}{terrestrial laser scanning}
\newacronym{UAV}{UAV}{unmanned aerial vehicle}
\newacronym{HD}{HD}{high definition}
\newacronym{RANSAC}{RANSAC}{RANdom SAmple Consensus}
\newacronym{ROI}{ROI}{region of interest}
\newacronym{DEM}{DEM}{digital elevation model}
\newacronym{ICP}{ICP}{iterative closest point}
\newacronym{SfM}{SfM}{structure from motion}
\newacronym{FME}{FME}{Feature Manipulation Engine}
\newacronym{OSM}{OSM}{OpenStreetMap} 
\newacronym{RMSE}{RMSE}{root mean square error}
\newacronym{CPT}{CPT}{conditional probability table}
\newacronym{DST}{DST}{Dempster–Shafer theory}
\newacronym{BN}{BayNet}{Bayesian network}
\newacronym{GIS}{GIS}{Geographic Information System}
\newacronym{PPD}{PPD}{posterior probability distribution}
\newacronym{CI}{CI}{confidence interval}
\newacronym{BoW}{BoW}{Bag of words}
\newacronym{HOG}{HOG}{Histogram of Oriented Gradients}
\newacronym{ORB}{ORB}{Oriented FAST and Rotated BRIEF}
\newacronym{BoVW}{BoVW}{bag-of-visual-words}
\newacronym{LGFBoVW}{LGFBoVW}{local–global feature BoVW}
\newacronym{VGI}{VGI}{ volunteered geographic information}
\newacronym{CNN}{CNN}{convolutional neural network}
\newacronym{BRIEF}{BRIEF}{Binary Robust Independent Features}
\newacronym{SIFT}{SIFT}{Scale Invariant Feature Transform}
\newacronym{USIP}{USIP}{unsupervised stable interest point detection}
\newacronym{OA}{OA}{overall accuracy}
\newacronym{PA}{PA}{producer's accuracy}
\newacronym{UA}{UA}{user's accuracy}

\begin{document}
\title{Reconstructing façade details using MLS point clouds and Bag-of-Words approach}
\titlerunning{Reconstructing façade details}
%
\author{Thomas Froech\textsuperscript{1 }, Olaf Wysocki\textsuperscript{1 }, Ludwig Hoegner\textsuperscript{1,2 }, Uwe Stilla\textsuperscript{1 }}
\authorrunning{Froech T., et al.,}
%
 \institute{Photogrammetry and Remote Sensing, TUM School of Engineering and Design, Technical University of Munich (TUM),\\ Munich, Germany - (thomas.froech, olaf.wysocki, ludwig.hoegner, stilla@tum.de) \and Department of Geoinformatics, University of Applied Science (HM), Munich, Germany - ludwig.hoegner@hm.edu\\}

%
%
\maketitle            
%
\begin{abstract}

In the reconstruction of façade elements, the identification of specific object types remains challenging and is often circumvented by rectangularity assumptions or the use of bounding boxes.
We propose a new approach for the reconstruction of 3D façade details. 
We combine \gls{MLS} point clouds and a pre-defined 3D model library using a \gls{BoW} concept, which we augment by incorporating semi-global features. 
We conduct experiments on the models superimposed with random noise and on the TUM-FAÇADE dataset \cite{Wysocki2023a}.
Our method demonstrates promising results, improving the conventional \gls{BoW} approach. It holds the potential to be utilized for more realistic facade reconstruction without rectangularity assumptions, which can be used in applications such as testing automated driving functions or estimating façade solar potential.

\keywords{point clouds \and façade reconstruction \and Bag-of-Words Approach \and mobile laser scanning}
\end{abstract}

\section{Introduction}

\begin{figure}[pt]
\centering
\includegraphics[width=0.9\textwidth]{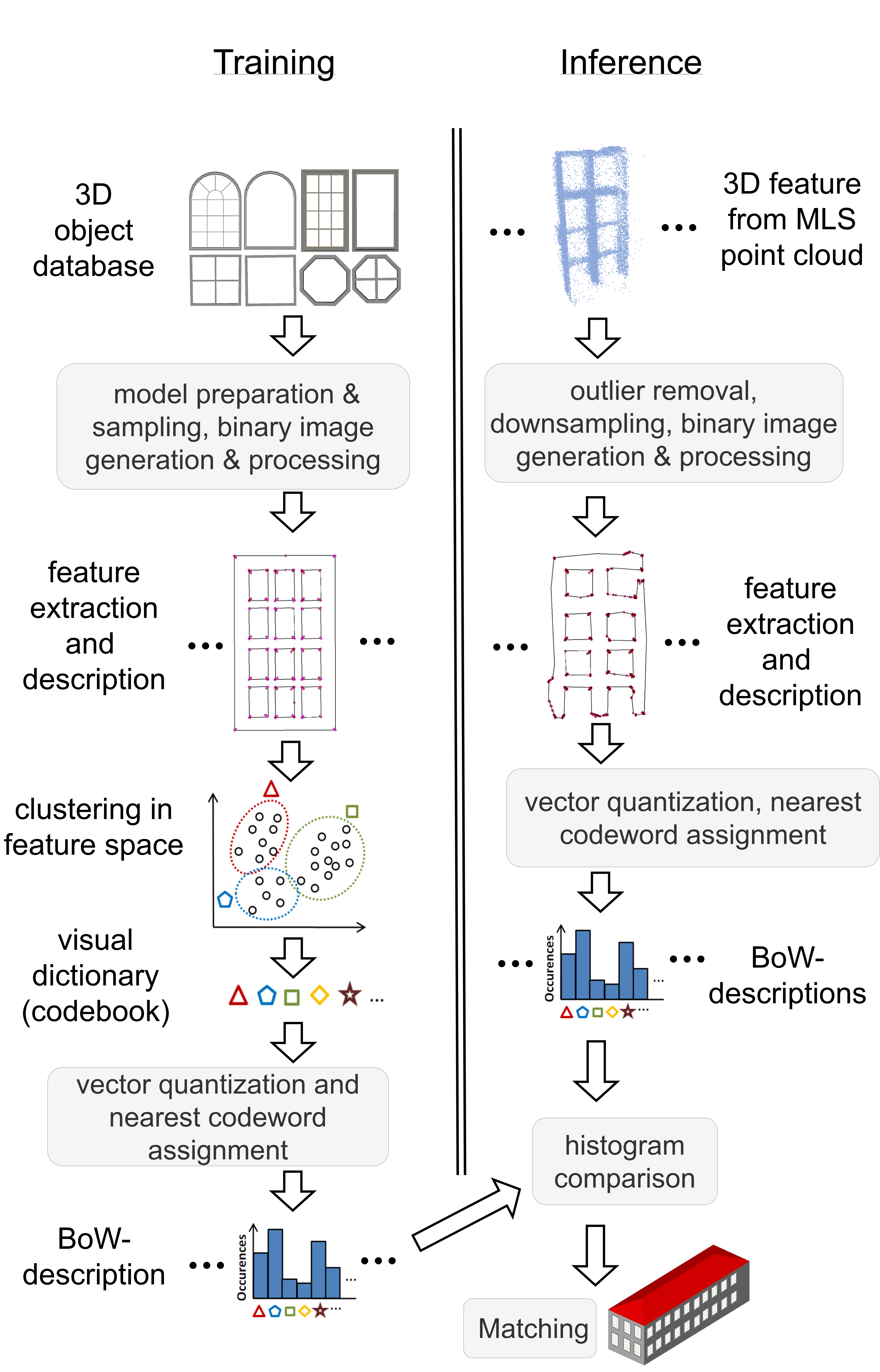}
\caption{Overview of the approach, left side: training process, right side inference(illustration based on~\cite{Memon2019})} \label{2eq:Overview}
\end{figure}

Semantic 3D building models up to \gls{LoD}2 level are widely used and available today~\cite{biljeckiApplications3DCity2015}.
~\gls{LoD}3 models characterized by a higher level of detail of their façade representation are scarce~\footnote{https://github.com/OloOcki/awesome-citygml}.
These are necessary for various applications, such as testing automated driving functions or estimating façade solar potential~\cite{Wysocki2023}.
The primary challenges in developing 3D façade reconstruction methods lie in the availability of street-level measurements and complexity of detailed façade element reconstruction; which is often circumvented by reconstructing the elements as bounding boxes~\cite{Hensel2019}.
Recently, however, there has been an increase in the availability of high-accuracy, street-level~\gls{MLS} point cloud data~\cite{Wysocki2023a}. 
In addition, databases that contain high-quality hand-modeled 3D façade details already exist~\cite{Trimble2021}.
These possess the potential to bridge the data-gap and move beyond the assumption of rectangularity and the use of bounding boxes.

In this paper, we propose an approach for the reconstruction of façade details leveraging the accuracy of~\gls{MLS} point clouds and the ubiquity of high quality 3D façade elements' libraries.
Specifically, we employ an enhanced bag-of-words (BoW) approach \cite{Csurka2004} to match measured façade elements with those from the library, without the rectangular assumptions.

\section{Related Work}
\subsection{\gls{LoD}3 Building Model Reconstruction}
The reconstruction of \gls{LoD}3 building models has attracted attention over an extended period of time \cite{helmutMayerLoD3,tuttas_reconstruction_2013,pantoja2022generating,ripperda_rekonstruktion_2010,Wysocki2023}.
Recent advancements have shown that façade elements reconstruction can be robustly performed, and yet identifying a specific object type remains challenging, for example, distinguishing between rectangular and oval window types~\cite{Wysocki2023}.
An example is the study of Hoegner and Gleixner which aims at the extraction of a 3D model of façades and windows from a \gls{MLS} point cloud\cite{Hoegner2022}. 
Their approach is based on a voxel octree structure and visibility analysis.
While they report a detection rate of $86\%$, they simplify windows and façades by representing them solely as rectangular shapes.

Other studies, such as that of Stilla and Tuttas, are devoted to the use of \gls{ALS} point clouds for the reconstruction of 3D building models \cite{tuttas_reconstruction_2013}.
They introduce an approach for the generation of façade planes with windows and the enrichment of a semantic city model with windows from a multi-aspect oblique view \gls{ALS} point cloud.

Following a different, image and deep learning-based approach for 3D model reconstruction, Fan \textit{et al.} propose VGI3D, an interactive platform for low-cost 3D building modeling from \gls{VGI} data using \gls{CNN}s in 2021 \cite{Fan2021}.
Their easy to use, lightweight, and quick application takes a small number images and a user's sketch of the façade boundary as an input \cite{Fan2021}.
For the automatic detection of façade elements, the object detection \gls{CNN} YOLO v3 \cite{Redmon2018} is utilized \cite{Fan2021}.

\subsection{Bag of Words Approach}
We use an adapted variant of the \gls{BoW} concept in our study.
The original \gls{BoW} approach is introduced by Salton and McGill in 1986 \cite{Salton1986}.
Csurka \textit{et al.} apply the~\gls{BoW} concept to images, where the detection and description of keypoints is one of the foundations of their approach:~\gls{BoVW}~\cite{Csurka2004}.
This concept is also applied to point cloud data.
Based on the Work of Xu \textit{et al.} on object classification of aerial images \cite{Xu2010}, Kang and Yang make use of a \gls{BoVW} approach to construct a supervoxel representation of raw point cloud data \cite{Kang2018}.
The \gls{BoVW} concept is further developed and adapted.
Zhu \textit{et al.} introduce the~\gls{LGFBoVW}.
They combine local and global features on histogram level.  
Such a combination increases the robustness compared to the traditional~\gls{BoVW}-approach~\cite{Zhu2016}.

\section{Methodology}
\subsection{Overview}

Figure \ref{2eq:Overview} provides an overview of our proposed method.
The training starts with 3D model preparation and sampling, where we create binary images from the sampled point clouds.
From these images, we extract and describe features, which are then clustered to obtain a visual dictionary.
By quantizing with the Euclidean distance, we assign the closest codeword to every feature vector.
Next, we count each codewords occurrences, providing the representations of the models as bags of codewords.
During the inference, we represent the target point clouds by the codebook.
To compare both representations, we employ histogram distances, where the model with the closest histogram distance to a target point cloud is selected as the best match.

\subsection{Sampling of the CAD models}
\begin{figure}[pt]
    \centering
    \includegraphics[width=0.5\textwidth]{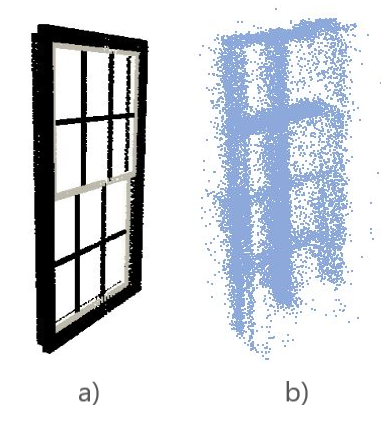}
    \caption{a) Point cloud sampled from CAD b) Point cloud extracted from the TUM-Façade dataset}
    \label{fig:Point_Cloud_Comparison}
\end{figure}

A key element of the method we propose is the correspondence between features extracted from the \gls{MLS} point clouds and the CAD models. 
In order to improve the comparability, we sample the CAD models with a sampling distance d to obtain a point cloud.
Figure \ref{fig:Point_Cloud_Comparison} a) gives an example of such a point cloud.
To enable comparison to point clouds from \gls{MLS} data, Figure \ref{fig:Point_Cloud_Comparison} b) shows an exemplary window that has been extracted from the TUM-Façade dataset.

The diversity of the CAD models does not allow for the same sampling distance to be used for all models.
The consequence of this could be sparse point clouds or point clouds with too many points.
Therefore, we choose d individually for the CAD models to obtain a suitable point cloud.
An alternative approach could be to normalize the CAD models before the sampling.
 gives an example of a point cloud sampled from a CAD model.

\subsection{Binary Image Generation and Processing}

We normalize the point clouds after outlier removal and downsampling.
To account for the decreasing point density with increasing height, we use an outlier removal that depends on the average height of the windows from the \gls{MLS} point cloud.

\begin{figure}[pt]
\centering
\includegraphics[width=0.75\textwidth]{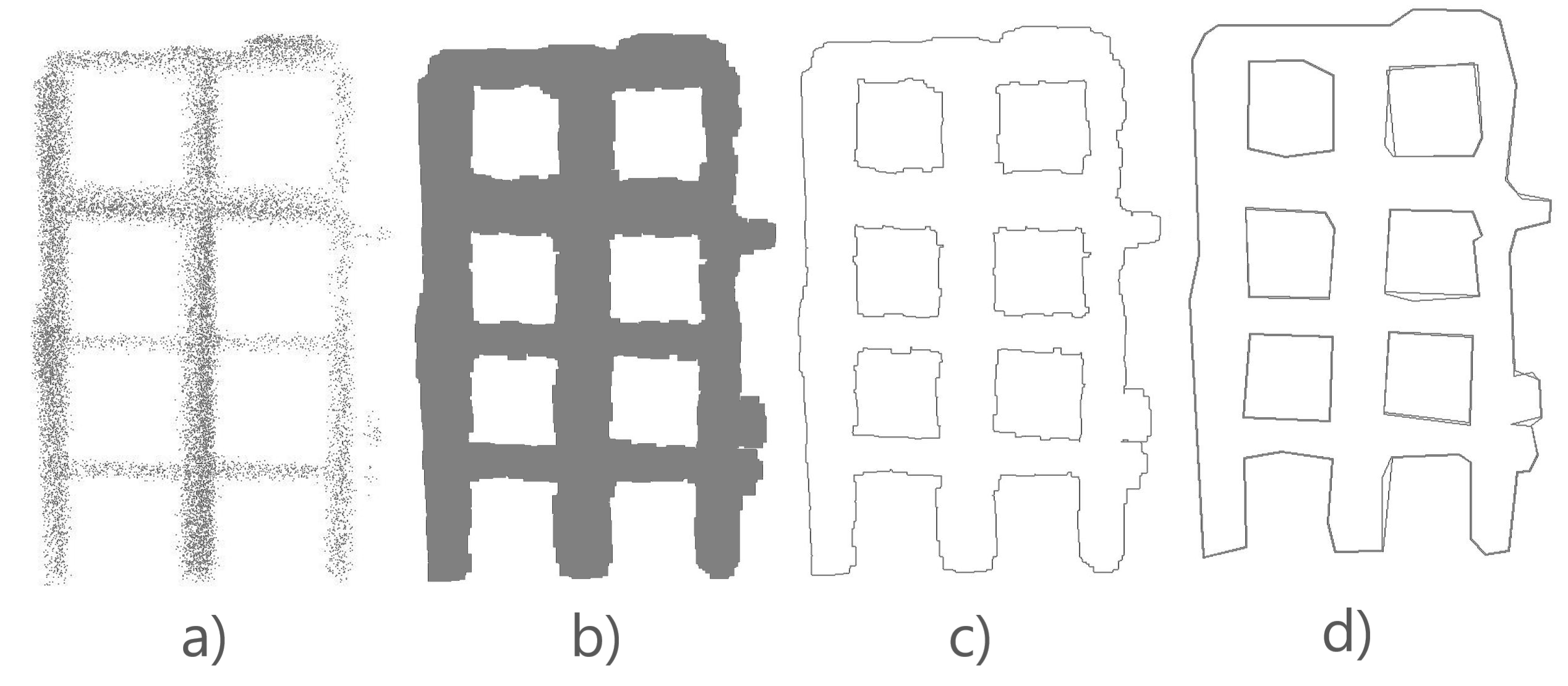}
\caption{Exemplary image processing on a projected point cloud: a) projected image of a point cloud, b) dilated image, c) edge detection (Laplace), d) line simplification (Douglas-Peucker).} \label{2eq:image_processing}
\end{figure}

\begin{figure}[pt]
\centering
\includegraphics[width=0.5\textwidth]{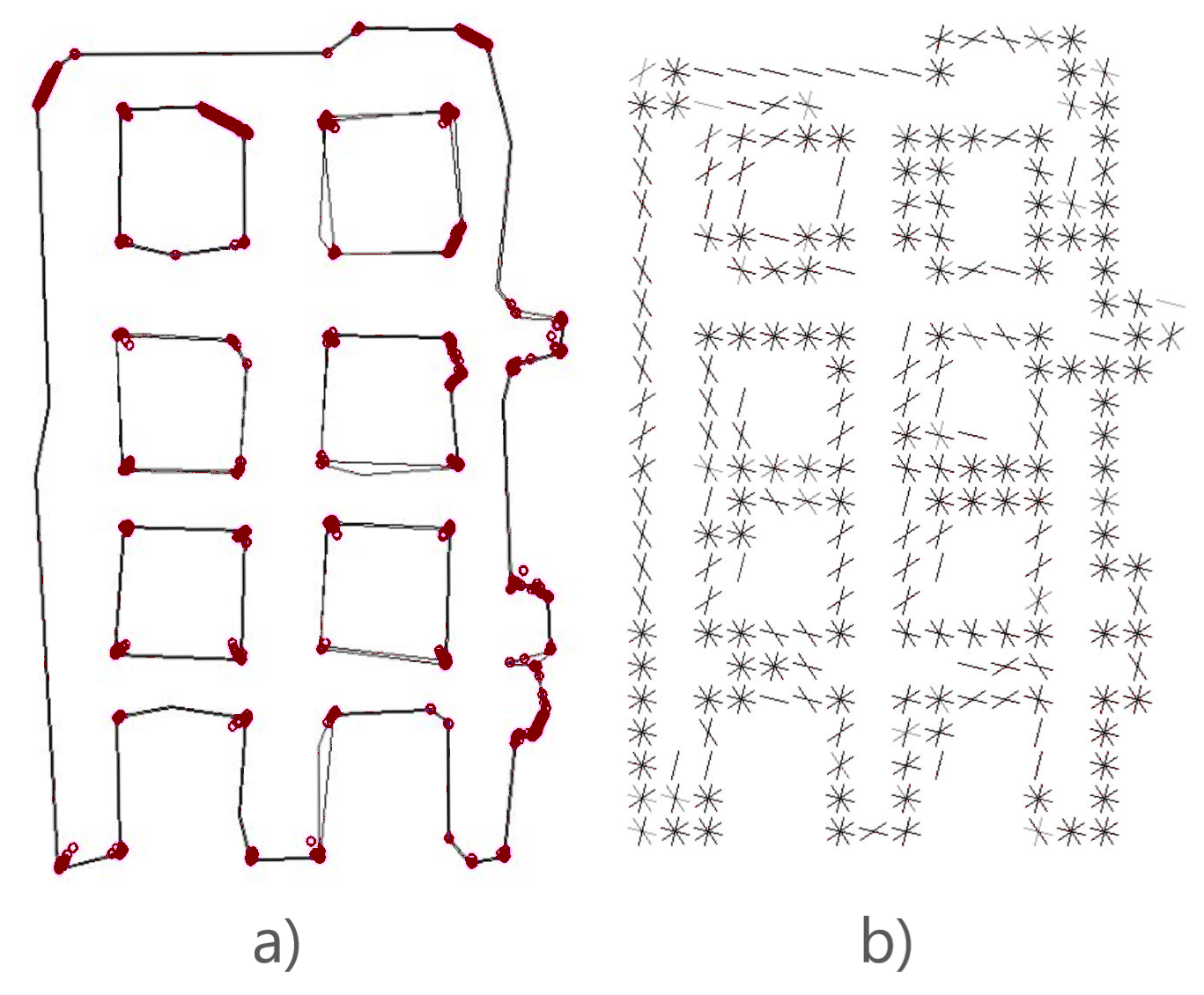}
\caption{Examples for feature extraction: a) ORB-keypoints b) HOG-image} \label{2eq:Hog_ORB_Example}
\end{figure}

Next, the point clouds are ortho-projected to a binary image ensuring its frontal view. 
As illustrated in Figure \ref{2eq:image_processing}, to enhance the extraction of meaningful keypoints, we apply the standard image processing techniques.
Figure \ref{2eq:Hog_ORB_Example} a) shows that the majority of the identified keypoints are located at semantically meaningful positions.

\subsection{Feature Extraction}

We use the \gls{ORB} \cite{Rublee2011} descriptor as a keypoint point detector.
This binary descriptor is based on the \gls{BRIEF} descriptor \cite{Rublee2011,Calonder2012}.
We choose this descriptor due to its resistance to noise and higher computational speed compared to other descriptors such as \gls{SIFT} \cite{Rublee2011,Lowe1999}.

In our study, we use dense feature sampling as an alternative approach to the interest point detection.
This method is opposed to the concept of identifying and describing key points.
In dense feature sampling, descriptors are sampled at points on a dense grid over the image, hence the name.
We sample the \gls{ORB} descriptor for each of the points in the dense grid.
This approach allows the extraction of a large amount of information at the cost of higher computational intensity \cite{Nowak2008}.

We use the \gls{HOG} feature descriptor \cite{Dalal2005} to incorporate semi-global information into the \gls{BoW} approach.
The fundamental concept of this descriptor is the investigation of the gradients and their orientation within the image on a dense grid \cite{Dalal2005}.
Normalized histograms of these gradients are established for each of these grid cells \cite{Dalal2005}.
The resulting one-dimensional vector characterizes the structure of the objects in the image. \cite{Dalal2005,Mallick2016}.

\subsection{Incorporation of Semi-Global Features}
\begin{figure}[pt]
\centering
\includegraphics[width=0.5\textwidth]{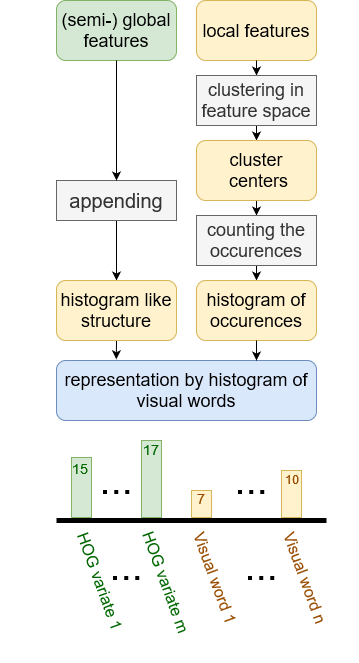}
\caption{Incorporation of semi-global features} \label{2eq:Incorporation_of_Global_Features}
\end{figure}

Generally, shapes possess a limited number of features~\cite{Bronstein2011}.
This poses difficulties in extracting large numbers of distinct features.
Semi-global information can be used to mitigate the effects of this issue.
However, semi-global features, as well as global features, cannot be directly integrated into the standard \gls{BoW} method.
Their (semi-) global uniqueness prevents the establishment of a frequency of occurence.
We propose a method to incorporate \gls{HOG} descriptors as semi-global information into the \gls{BoW}-approach to overcome this issue.
Figure \ref{2eq:Hog_ORB_Example} b) gives an example of the information obtained with \gls{HOG}.
A 2D diagram that displays the distribution of the gradients in the respective cell is shown for every cell in the image, in which a gradient is present.
Figure \ref{2eq:Incorporation_of_Global_Features} gives an overview of our concept.
We establish a structure similar to a histogram by considering each \gls{HOG} descriptor variate as a separate histogram bin, with the value of the bin equaling the value of the corresponding variate.
This structure comprises as many bins as \gls{HOG} variates.
We concatenate the occurence histogram that we obtain from the \gls{BoW} approach with the histogram-similar structure constructed from the HOG-variates to a combined histogram.

\subsection{Clustering}
For simplicity, we make use of the standard K-Means clustering algorithm in our \gls{BoW} approach.
We treat every descriptor variate as an axis in this clustering.
Our clustering problem thus has the same dimensionality as the extracted feature descriptor.
With K-Means clustering, the number of clusters n has to be determined in advance.
The setting of this hyper-parameter is critical for the performance of the \gls{BoW} method since the meaningful assignment of data points to cluster centers depends on it \cite{Kang2018}.
There appear only a few empty or overcrowded cluster centers for this n.
We acknowledge that there might be better configurations for n, and that there are more sophisticated clustering algorithms available that might lead to improved results.

\subsection{Histogram Comparison}
We use histogram distances to assess the similarity of the histograms within our \gls{BoW} approach.
There is a large number of different histogram distances available \cite{Cha2008}.
The specific choice of histogram distance employed holds limited significance regarding the general approach that we introduce, thereby allowing for arbitrary selection in our methodology.
We make use of the histogram distances that are introduced in the following

The Minkowski distance is given as \cite{Cha2008}:

\begin{equation}
D = \left(\sum_{i=1}^{n} \left|q_i - p_i\right|^p\right)^{\frac{1}{p}}
\end{equation}

The Minkowski distance is characterized by the parameter p.
It can be interpreted as a more general form of the Manhattan distance $(p=1)$, the euclidean distance $(p=2)$ and the Chebyshev distance $(p=\infty)$ \cite{Cha2008}.
It can be used to evaluate the similarity of two histograms by pairwise comparison of the individual bins and subsequent accumulation of the obtained distances.

The Jensen-Shannon divergence is given as \cite{Bamler2022}:

\begin{equation}
\text{{JSD}}(P, Q) = \frac{1}{2} \left( \text{{KL}}(P \| M) + \text{{KL}}(Q \| M) \right)
\end{equation}

where $\text{{KL}}(P \| Q)$ is the Kullback-Leibler-Divergence between the probability distributions $P$ and $Q$.
The Kullback-Leibler-Divergence given as \cite{Bamler2022}:

\begin{equation}
\text{{KL}}(P \| Q) = \sum_{i} P(i) \ln \left( \frac{P(i)}{Q(i)} \right)
\end{equation}

The Jensen-Shannon divergence, like the Kullback-Leibler divergence, is based on the concept of entropy \cite{Cha2008} and is used to quantify the similarity of two probability distributions. 
The Jensen-Shannon divergence can be interpreted as a symmetric version of the Kullback-Leibler divergence \cite{Cha2008}.

The Pearson Chi-Square-Distance between the probability distributions $P$ and $Q$ is given as \cite{Cha2008}:

\begin{equation}
D\chi^2(P, Q) = \sum_{i} \frac{(P(i) - Q(i))^2}{P(i)}
\end{equation}

\section{Experiments}
We infered our method on the models, superimposed with random noise and on a building from the labeled~\gls{MLS} point cloud of the TUM-FAÇADE data set~\cite{Wysocki2023a}.
To ensure the comparability of our experiments, we consistently used the same number of clusters n in all of them.

We found a suitable n in a heuristic way.
We performed experimental clustering and feature descriptor quantization for different values of n.
We then evaluated the frequency of occurrence of the respective cluster centers.
We find that in most of our cases 25 clusters give satisfactory results.

\begin{figure}[pt]
\includegraphics[width=\textwidth]{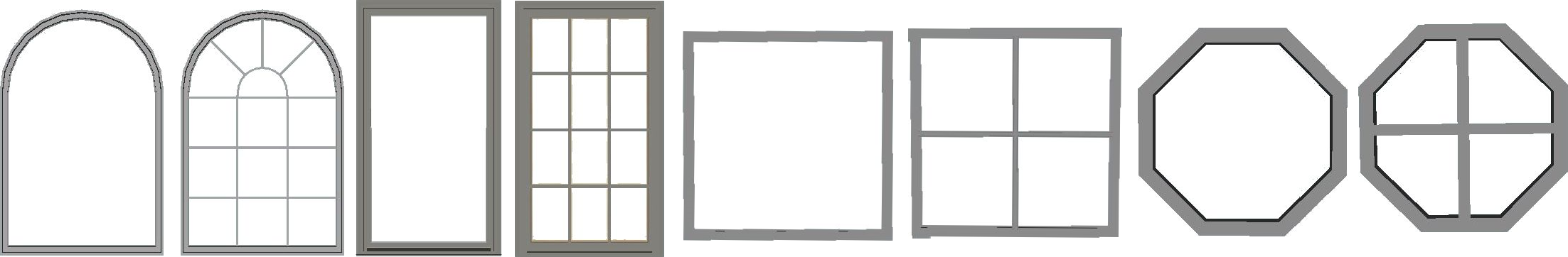}
\caption{3D window CAD models} \label{2eq:all_models}
\end{figure}

We acquired four CAD models from the SketchUp 3D Warehouse library~\cite{Trimble2021} (Figure \ref{2eq:all_models}).
We chose the models so that each window that is present in the TUM-Façade dataset at least matches one of the models.
We additionally chose one window that is of octagon like shape as a model without a matching window in the dataset.We manually edited some of the CAD models to add or remove window bars.

\begin{figure}[pt]
    \centering
    \includegraphics[width=0.45\textwidth]{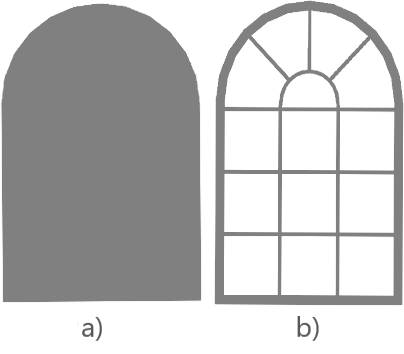}
    \caption{Point clouds, sampled from CAD model: a) without removal of window glass b) with removal of window glass }
    \label{fig:glass_removal}
\end{figure}

When sampling the CAD-models to a point cloud, transparency properties have to be taken into account.
As shown in Figure \ref{fig:glass_removal}, we removed glass from all models to exploit the presence or absence of window bars.
In Figure \ref{fig:glass_removal} b) detailed window bars become apparent due to glass components not being sampled.
In contrast, Figure \ref{fig:glass_removal} a) illustrates that when the glass components are sampled using the same method as the rest of the window, it results in the window appearing as an opaque object.
With MLS scans, the laser beams usually penetrate the glass and thus penetrate into the interior of the building.
This justifies the removal of glass from the CAD models.

We evaluate our results by determining the overall accuracy, the user's accuracy the producer's accuracy, and Cohen's kappa coefficient.

The \gls{OA} is calculated by dividing the number of correctly classified pixels by the total number of pixels \cite{Congalton2019}:

\begin{equation}
\text{{OA}} = \frac{\text{{number of correctly classified samples}}}{\text{{total sample number}}} 
\end{equation}

The \gls{PA} is used to determine the percentage of the ground truth samples that is classified correctly \cite{Congalton2019}:

\begin{equation}
\text{{PA}} = \frac{\text{{number of correctly classified samples of a class X}}}{\text{{total number of ground truth samples in class X}}} 
\end{equation}

The \gls{UA} is used to calculate the percentage of the classification result of a class that is classified correctly \cite{Congalton2019}:

\begin{equation}
\text{{UA}} = \frac{\text{{number of correctly classified samples of a class X}}}{\text{{total number of samples classified as class X}}} 
\end{equation}

Cohen's kappa coefficient represents an alternative to using \gls{OA}.
It has a value range from 0 to 1, where 0 represents complete randomness while 1 would represent a perfect classifier \cite{Congalton2019}.
It can be interpreted as a measure for the concordance between the predicted class assignments and the class assignment of the ground truth data \cite{Grandini2020}.
It can be calculated from the confusion matrix according to the following formula:

\begin{equation}
\kappa = \frac{\text{{OA}}-\text{{RM}}}{1-\text{{RM}}}
\end{equation}

with the random match (RM):

\begin{equation}
\text{{RM}} = \frac{\sum(\text{{product of row and column sums}})}{(\text{{total sum}})^2}
\end{equation}

The implementation is available in the repository \cite{froech2023}.

\subsection{Experiments on models superimposed with random noise}
\begin{figure}[pt]
    \centering
    \includegraphics[width=0.75\textwidth]{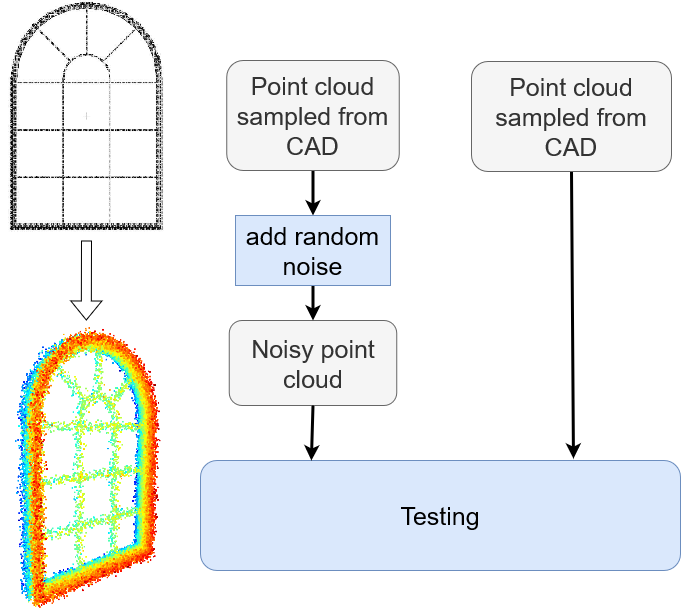}
    \caption{Schema: Addition of random noise to the point clouds sampled from CAD}
    \label{fig:noisy_cloud}
\end{figure}

\begin{figure}[pt]
\includegraphics[width=\textwidth]{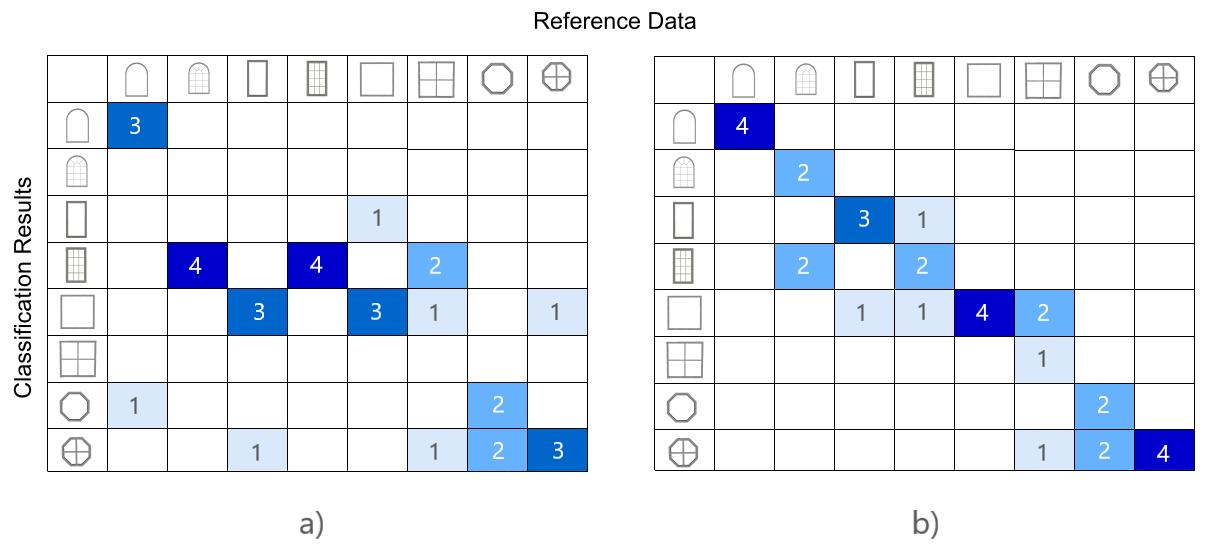}
\caption{Confusion matrices: a) ORB, b) ORB and HOG} \label{2eq:Small_Scale_Experiments}
\end{figure}

\begin{figure}[pt]
\centering
\includegraphics[width=0.5\textwidth]{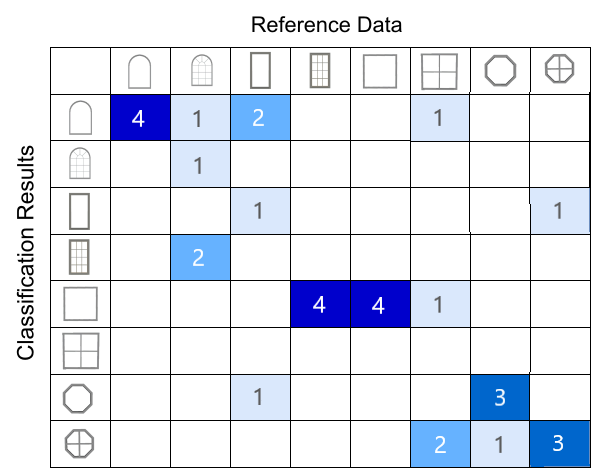}
\caption{Confusion matrix: ORB and HOG, noise level with factor 2} \label{2eq:Confusion_Matrix_04}
\end{figure}

We added random noise to the point clouds that we sample from the pre-processed CAD models, as schematically described in Figure \ref{fig:noisy_cloud}.
We used the Chi-Square histogram distance for these experiments.
Results from these experiments are summarized in Figure \ref{2eq:Small_Scale_Experiments}, Figure \ref{2eq:Confusion_Matrix_04} and Table \ref{tab:table1}.

We find that incorporating~\gls{HOG} descriptors improves the matching quality.
Table \ref{tab:table1} shows that the \gls{OA} is improved from $0.47$ to $0.69$ this way.
The kappa-coefficient is improved from $0.36$ to $0.65$.
We observe sensitivity towards noise.
By doubling the noise level, the kappa coefficient is diminished to $0.43$, while the overall \gls{OA} drops to $0.5$.

In addition, we identify a dependency of the matching quality on the type of CAD model.
Figures \ref{2eq:Small_Scale_Experiments} and \ref{2eq:Confusion_Matrix_04} show that the matching is more stable for the arched and octagon shaped windows than for the rectangular and quadratic windows.

\begin{table}[pt]
\centering
\caption{Results of the experiments with the models superimposed with random noise}
\label{tab:table1}
\begin{tabular}{|l|l|l|}
\hline
Experiment &  Overall Accuracy & Kappa Coefficient\\
\hline
ORB, Chi-Square dist. &  0.47 & 0.36\\
ORB and HOG, Chi-Square dist. &  0.69 & 0.65\\
ORB and HOG, Chi-Square dist., noise $*2$ & 0.50 & 0.43\\
\hline
\end{tabular}
\end{table}

\begin{figure}[pt]
    \centering
    \includegraphics[width=\textwidth]{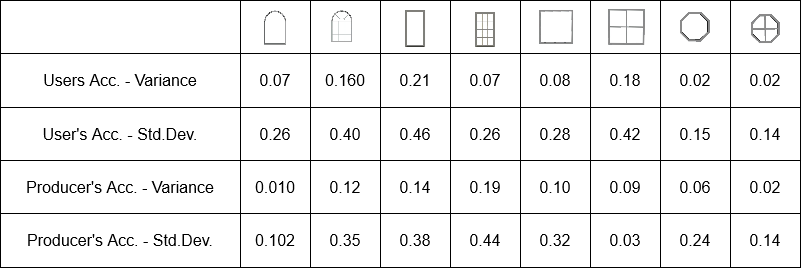}
    \caption{Standard deviation and variance of user's and producer's accuracies from experiments with six different combinations of hyper parameters.}
    \label{fig:analysis}
\end{figure}

We conducted further experiments with different combinations of features and histogram distances.
More information on the experiments can be found here.
Figure \ref{fig:analysis} summarizes the standard deviations and the variances of the user's and producer's accuracies for six of these experiments with different sets of hyper parameters.
We find that the arched window with no bars and the two octagon-shaped windows are matched in the most stable manner compared to the other window types.

\subsection{Experiments on the TUM-FAÇADE dataset}

\begin{figure}[pt]
\begin{center}
\includegraphics[width=0.75\linewidth]{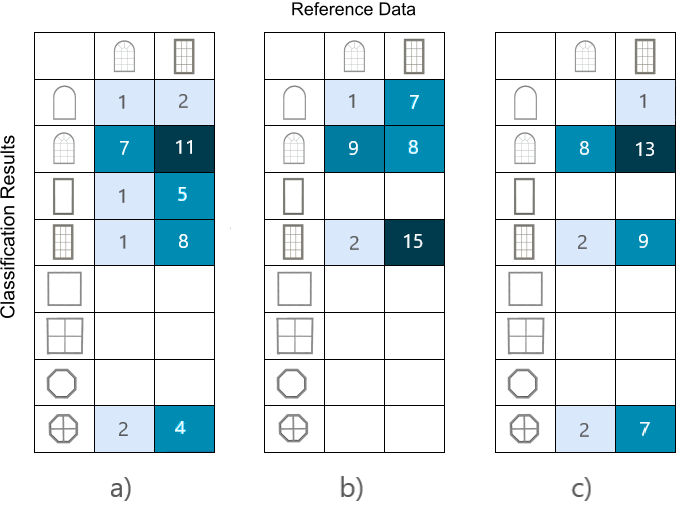}
\end{center}
\caption{a) ORB, Jensen-Shannon Divergence, b) ORB and HOG, Jensen-Shannon Divergence, c) ORB and HOG, Minkowski-Distance}
\label{2eq:eight}
\end{figure}

\begin{figure}[pt]
\centering
\includegraphics[width=\textwidth]{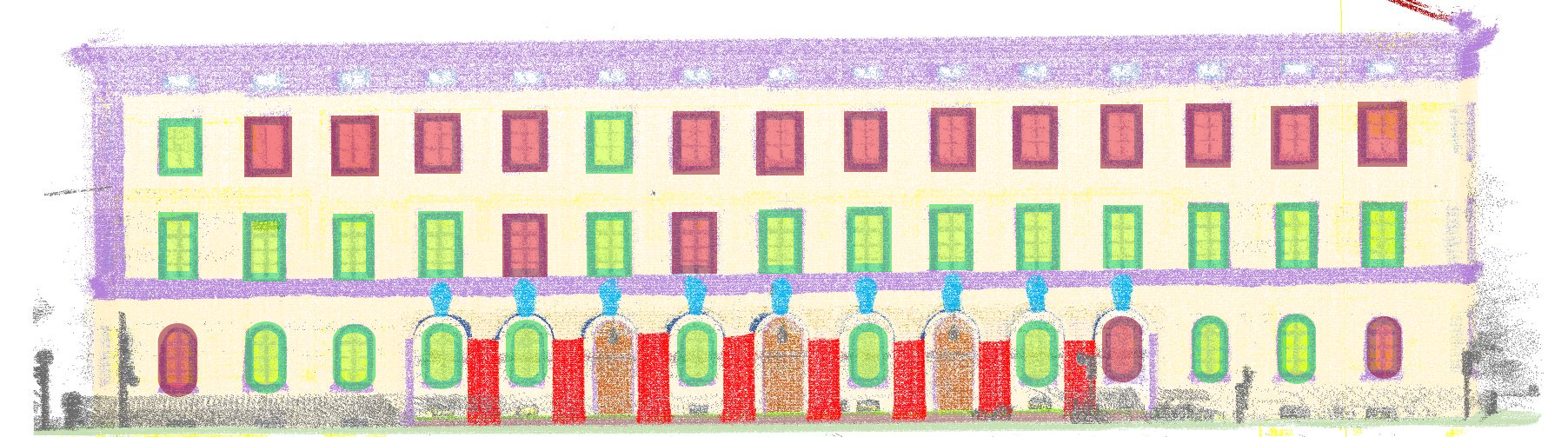}
\caption{Correctly (green) and falsely (red) matched windows} \label{2eq:TUM_Facade_Performance.}
\end{figure}

The tested façade only comprises rectangular and arched windows with window bars (Figure \ref{2eq:TUM_Facade_Performance.}).
We observe an improvement of the matching quality with the incorporation of~\gls{HOG}-descriptors, as the \gls{OA} increases to $0.57$.
Also, the dependence on the histogram distance is shown in Table \ref{tab:table2}.

\begin{table}[pt]
\centering
\caption{Results of the experiments on the TUM-FAÇADE dataset \cite{Wysocki2023a}}\label{tab:table2}
\begin{tabular}{|l|l|}
\hline
Experiment &  Overall Accuracy \\
\hline
ORB, Jensen-Shannon Divergence &  0.36 \\
ORB and HOG, Jensen-Shannon Divergence &  0.57 \\
ORB and HOG, Minkowski-Distance & 0.41\\
\hline
\end{tabular}
\end{table}

\begin{figure}[pt]
    \centering
    \includegraphics[width=0.35\textwidth]{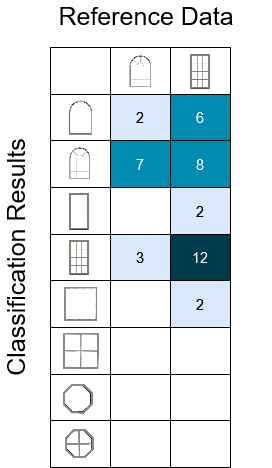}
    \caption{Results of the experiments on the TUM-FAÇADE dataset with dense feature sampling}
    \label{fig:conf_mat_dfs}
\end{figure}

We observe a correlation of the matching quality with the building height in Figure \ref{2eq:TUM_Facade_Performance.}.
This can presumably be attributed to the point density of the~\gls{MLS} point cloud, which decreases with increasing altitude.

The experiment with \gls{ORB}, \gls{HOG}, and the Jensen-Shannon Divergence was also conducted using dense feature sampling instead of extracting key points.
Figure \ref{fig:conf_mat_dfs} summarizes the results from this experiment.
However, with an \gls{OA} of $0.45$, we observe a decrease in in accuracy.

\section{Discussion}

\subsection{3D interest point detection}
In our study, we employ 2D projections of point clouds to generate binary images and extract different types of features from them.
We adopt this approach based on the intuition that the front view contains most of the information that characterizes the window. 

The point cloud data as well as the CAD models are three-dimensional.
Therefore, instead of a projection and subsequent extraction of 2D feature descriptors, it seems obvious to use 3D interest point detection directly.
There is a number of such operators available.
Examples are \gls{USIP} for points clouds \cite{Li2019} or Harris 3D for meshes \cite{Sipiran2011}.

Assuming that the front view of a window contains the majority of information that characterizes a specific window type, the potential benefit derived from incorporating 3D information is expected to be marginal.
The desired correspondence between the point cloud that is sampled from the CAD model and the \gls{MLS} point cloud represents an obstacle when using such detectors.
The often irregular 3D shape of the windows from the MLS point clouds does not necessarily correspond to the very regular 3D-shape of the windows that are sampled from the CAD models. 
Therefore, the extracted keypoints are most probably not going to correspond to the keypoints found in the model in most cases.

\subsection{Influences of the Histogram Distance}
We observe, that the use of the chi-square histogram distance yields the best results in the set of experiments on the models superimposed with random noise (see Section 4.1).
However, due to the asymmetry of the particular histogram distance used, these results should be viewed with caution.
The asymmetry could lead to unbalanced assessments of the similarity of histograms.
Using the symmetrical form of the chi-square histogram distance \cite{Cha2008} may yield more meaningful results in the context of this study.
In the experiments with the TUM-FAÇADE dataset, the use of the Jensen-Shannon divergence leads to the best results.
This histogram, distance is, in contrast to the Chi-Square distance that we use, symmetric.

\subsection{Dense feature sampling}
We find that using dense feature samples does not improve the performance of our method compared to the other methods.
We attribute this to the binary properties of the images that we generate from the projected point clouds.
Figure \ref{2eq:Hog_ORB_Example} a) demonstrates that almost all important interest points that characterize the object in the image are identified by the \gls{ORB} keypoint detector.
When using dense feature sampling, most points for which the descriptors are determined are located in empty regions of our binary images or in the vicinity of straight lines.
Therefore the information gain of using dense feature sampling is not particularly large in context of our method.

\section{Conclusion}
In this paper we present a method for reconstructing façade details by using \gls{MLS} point clouds and the~\gls{BoW} concept. 
We incorporate semi-global features to address the issue of insufficient distinct features in shapes.
In our two sets of experiments, we demonstrate improved performance compared to the conventional~\gls{BoW} approach.
Our method seems to be sensitive to noise and point cloud sparsity.
Future work could focus at increasing the accuracy of the method as well as its computational efficiency.
In the Future, our method could be applied in façade reconstruction pipelines that aim at a more realistic reconstruction without assumptions of rectangularity or the use of bounding boxes.

\section{Acknowledgements}
This work was supported by the Bavarian State Ministry for Economic Affairs, Regional Development and Energy within the framework of the IuK Bayern project \textit{MoFa3D - Mobile Erfassung von Fassaden mittels 3D Punktwolken}, Grant No.\ IUK643/001.
Moreover, the work was conducted within the framework of the Leonhard Obermeyer Center at the Technical University of Munich (TUM).
We gratefully acknowledge the Geoinformatics team at the TUM for the valuable insights and for providing the CityGML datasets.

\bibliographystyle{splncs04}
\bibliography{bibliography}

\end{document}